\documentclass[conference]{IEEEtran}
\IEEEoverridecommandlockouts
\usepackage{cite}
\usepackage{amsmath,amssymb,amsfonts}
\usepackage{algorithmic}
\usepackage{graphicx}
\usepackage[none]{hyphenat}
\usepackage{ragged2e}
\usepackage{url}
\usepackage{booktabs}

\usepackage{textcomp}
\usepackage{xcolor}
\usepackage[linesnumbered,ruled,vlined]{algorithm2e}
\def\BibTeX{{\Large\rm B\kern-.05em{\sc i\kern-.025em b}\kern-.08em
    T\kern-.1667em\lower.7ex\hbox{E}\kern-.125emX}}
\begin{document}

\title{Curating Stopwords in Marathi: A TF-IDF Approach for Improved Text Analysis and Information Retrieval}

\author{\IEEEauthorblockN{Rohan Chavan}
\IEEEauthorblockA{\textit{Computer Science and Engineering} \\
\textit{Walchand College of Engineering, Sangli}\\
Sangli, Maharashtra, India \\
rohanchavank22@gmail.com}
\\
\IEEEauthorblockN{Gaurav Patil}
\IEEEauthorblockA{\textit{Computer Science and Engineering} \\
\textit{Walchand College of Engineering, Sangli}\\
Sangli, Maharashtra, India \\
gauravpatil1407@gmail.com}
\and
\IEEEauthorblockN{Vishal Madle}
\IEEEauthorblockA{\textit{Computer Science and Engineering} \\
\textit{Walchand College of Engineering, Sangli}\\
Sangli, Maharashtra, India \\
vishalmadle13@gmail.com}
\\
\IEEEauthorblockN{Raviraj Joshi}
\IEEEauthorblockA{\textit{L3Cube Labs, Pune} \\
\textit{Indian Institue of Technology, Madras}\\
Pune, Maharashtra, India \\
ravirajjoshi@gmail.com}
}

\maketitle

\begin{abstract}
Stopwords are commonly used words in a language that are often considered to be of little value in determining the meaning or significance of a document. These words occur frequently in most texts and don't provide much useful information for tasks like sentiment analysis and text classification. English, which is a high-resource language, takes advantage of the availability of stopwords, whereas low-resource Indian languages like Marathi are very limited, standardized, and can be used in available packages, but the number of available words in those packages is low. Our work targets the curation of stopwords in the Marathi language using the MahaCorpus, with 24.8 million sentences. We make use of the TF-IDF approach coupled with human evaluation to curate a strong stopword list of 400 words. We apply the stop word removal to the text classification task and show its efficacy. The work also presents a simple recipe for stopword curation in a low-resource language. The stopwords are integrated into the mahaNLP library and publicly available on \url{https://github.com/l3cube-pune/MarathiNLP}.
\end{abstract}
\vspace{12pt}
\begin{IEEEkeywords}
 Natural Language Processing, Stopwords Curation, Marathi Language, TF-IDF Approach, Text Preprocessing, Marathi Stopword List, Indic NLP.
\end{IEEEkeywords}

\section{\textbf{\Large Introduction}}
Stopwords, those frequently occurring yet semantically less informative words such as 'the', 'and', 'is', and 'in' in English, and words like '(\includegraphics[height=1em]{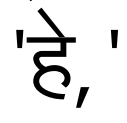})', '(\includegraphics[height=1em]{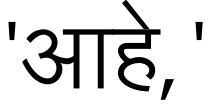})', and '(\includegraphics[height=1em]{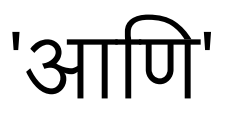})' in Marathi, are foundational elements in Natural Language Processing (NLP). They act as linguistic connectors, maintaining sentence structure and grammar, but are often filtered out during text analysis, as they typically contribute less to the extraction of meaningful insights from textual data. In the context of preprocessing in Natural Language Processing (NLP), stopword removal is one of the most common tasks since it helps reduce data noise, improve efficiency, and enhance information retrieval \cite{silva2003importance}.

\begin{algorithm}[]
\SetAlgoNlRelativeSize{-1}
\SetKwProg{Fn}{Function}{:}{}
\Fn{\texttt{RemoveStopwords}}{
    \KwData{Text input: \texttt{text}}
    \KwResult{Text with stopwords removed}
    \texttt{from mahaNLP.preprocess import Preprocess}\;
    \texttt{preprocessor = Preprocess()}\;
    \texttt{text = \includegraphics[height=1em]{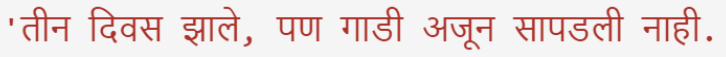}}
    \texttt{\includegraphics[height=1em]{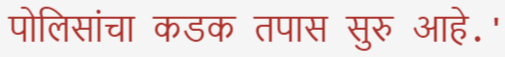}}\;
    \texttt{filtered\_text = preprocessor.remove\_stopwords(text)}\;
}
\caption{Using MahaNLP to Remove Stopwords}
\label{algo:algo_1}
\end{algorithm}

Algorithm \ref{algo:algo_1} illustrates the use of the MahaNLP library to remove stopwords in Marathi text. The code snippet showcases the simplicity of integrating stopword removal into the preprocessing pipeline, emphasizing the efficiency of the proposed approach.
\vspace{8pt}

The presence of curated stopword lists is commonplace for high-resource languages, facilitating accurate text analysis. However, the landscape shifts when we turn our attention to low-resource languages, such as Marathi. Despite being spoken by a significant population, Marathi lacks comprehensive computational resources, including a well-curated stopword list \cite{joshi2022l3cube_mahanlp}. This gap poses a substantial challenge for NLP in Marathi, impeding the effectiveness of text analysis tasks and limiting the development of Marathi language technology applications. The code snippet Algo. \ref{algo:algo_1} demonstrates how the functionality of removing stopwords can be carried out using the mahaNLP library Python script.
\vspace{8pt}

In this work, we present an approach to address these challenges by curating a stopword list in Marathi. Our methodology leverages the TF-IDF (Term Frequency-Inverse Document Frequency) technique, a well-established method in NLP for quantifying term importance within a corpus \cite{sparck1972statistical}. The key innovation of our approach lies in its usage of the MahaCorpus \cite{joshi2022l3cube}, a vast Marathi text corpus comprising 24.8 million sentences, meticulously curated by L3Cube. By utilising TF-IDF scores on this extensive dataset, we systematically identify and rank stopwords, creating a resource that captures the unique linguistic intricacies and content nuances specific to Marathi. We perform human evaluation in the end to curate strong gold standard stopwords comprising 400 words. We also evaluate the impact of stop-word filtering on the L3Cube-MahaNews text classification task to show that there is minimal impact on the downstream accuracy.

\begin{figure*}[!ht]
  \centering
  \includegraphics[width=0.9\textwidth]{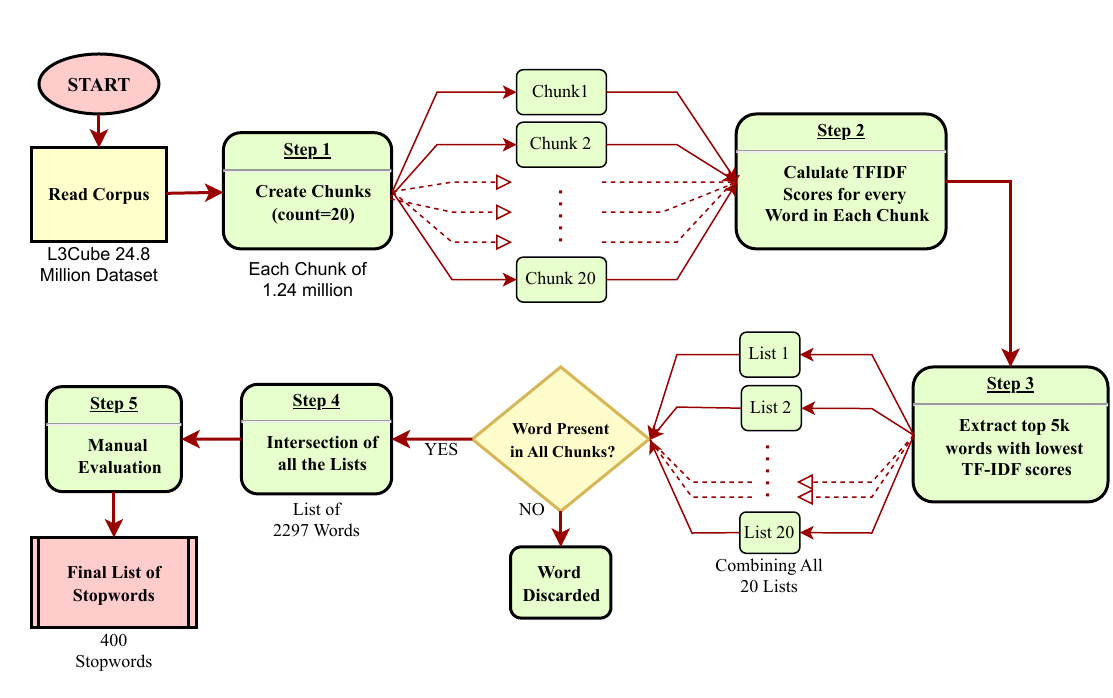}
  \caption{Visual representation of the methodology used for curation of stopwords}
  \label{Flow Diagram}
\end{figure*}

\vspace{8pt}

The main contributions of our work are as follows:
\vspace{8pt}

\begin{itemize}
    \item We present a simple hybrid approach for stop-word curation for a low-resource language. The approach utilizes both automatic and human judgements for stop words identification from a large unsupervised corpus.
    \item We present the largest stopwords list in Marathi with 400 words. It is available publicly and also accessible via the mahaNLP library \cite{magdum2023mahanlp}. 
\end{itemize}
\vspace{8pt}

Our focus on stopword removal aligns with the common preprocessing tasks in NLP, where it acts as a fundamental step to reduce data noise, enhance efficiency, and improve information retrieval \cite{silva2003importance}. While the presence of curated stopword lists is well-established for high-resource languages, the scarcity of such resources in low-resource languages, including Marathi, poses a significant challenge. This work addresses this gap by presenting a novel approach to curating a comprehensive stopword list for Marathi, utilizing TF-IDF on the extensive MahaCorpus. Our methodology, validated through human evaluation and downstream accuracy assessments, contributes to advancing Marathi language technology and NLP applications.
\vspace{8pt}

\section{\textbf{\Large Related Work}}
\vspace{8pt}
The curation of stopwords play a pivotal role in enhancing the efficiency and precision of various language based applications in the field of Natural Language Processing. Due to the unique linguistic characteristics of different languages, the significance of curation of stopwords for a particular language is immense as the stopwords need to be language-specific to be effective for improvement. Many studies have been done on the curation of language-specific stopwords in various different languages, this paper particularly focuses on the curation of stopwords for Marathi language which is a language spoken in the state of Maharashtra in India. Here is a review of the previous research efforts used for stopwords curation in different languages and their unique approach to curate a language-specific stopwords list best suited for that language.
\vspace{8pt}

Stopwords curation is a vital component of text processing and natural language processing (NLP) in multiple languages. Several studies have explored the creation and application of stopword lists in diverse linguistic contexts. For instance, Pandey and Siddiqui 2009 \cite{pandey2009evaluating} focused on the Hindi language, generating a list of stopwords based on frequency and manual analysis. They applied this list to improve text retrieval accuracy, highlighting the essential role of stopword removal in enhancing retrieval outcomes. Similarly, Kaur and Saini 2015\cite{kaur2015natural} concentrated on Punjabi literature, specifically poetry and news articles, to identify and release 256 stop words for public use.
Another study Kaur and Saini 2016\cite{kaur2016punjabi} published by them contains a  list which consists of stop words of Punjabi language with their Gurmukhi, Shahmukhi as well as Roman scripted forms.
\\ \\
Yao and Ze-Wen 2011\cite{yao2011research} compiled a comprehensive list of 1,289 Chinese-English stop words by combining domain-specific stop words with classical stop words, emphasizing the importance of considering various categories of stopwords. Alajmi et al. \cite{alajmi2012toward} employed a statistical approach to generate a stopword list for the Arabic language, demonstrating the applicability of statistical methods in diverse linguistic contexts.
Savoy \cite{savoy1999stemming} introduced a comprehensive list of general stop words tailored for French, addressing the necessity to filter out frequently used yet non-essential terms in document retrieval. They devised this stop word list using a three-fold approach inspired by Fox's \cite{fox1989stop} guidelines.
First being sorting all word forms in their French corpora by frequency, retaining the top 200 most common words. Next, they eliminated numbers, as well as nouns and adjectives loosely associated with the main subjects of the collections. Finally, they included various non-information-bearing words like personal pronouns, prepositions, and conjunctions, even if they weren't among the top 200. The resulting French general stop-word list comprises 215 words. Its application led to a remarkable reduction of approximately 21\% in the inverted file size for one test collection and about 35\% for a second corpus. 
The first including sorting by frequency, next elminating nouns, numbers and adjectives and finally including non-information bearing words like pronouns and conjunctions.
\\ \\
In the context of Sentiment Analysis, Ghag and Shah 2015 \cite{ghag2015comparative} reported improvements in classifier accuracy from 50\% to 58.6\% in sentiment analysis when stopwords were eliminated, emphasizing the relevance of stopwords in fine-tuning classification algorithms.
\\ \\
Rakholia and Saini 2017 \cite{rakholia2017rule} conducted a comprehensive analysis of different stemmer algorithms and pre-processing approaches for the Gujarati language. Their research emphasized the significance of stop words removal as a fundamental pre-processing step in natural language processing applications. Furthermore, they introduced a rule-based approach for dynamically identifying Gujarati stopwords, achieving impressive accuracy rates for generic and domain-specific stopword detection.
\\ \\
Recent research by Fayaza and Fathima Farhat 2021 \cite{fayaza2021towards} proposed a dynamic approach for Tamil stopword identification in text clustering. Their work underscores the significance of a dynamic approach to stopwords, emphasizing the need for language-specific strategies in text analysis and NLP.
In a different context, Siddiqui and Sharan 2018 \cite{siddiqi2018construction} delved into the construction of a generic stopword list for the Hindi language, without relying on corpus statistics. Their work highlights the domain-independent nature of stopwords and the potential of stopword lists to enhance various text analysis applications.
\vspace{8pt}

\section{\textbf{\Large Methodology}}
\vspace{8pt}

\subsection{\textbf{Data Collection and Preprocessing}}

Our Approach (Fig.1) begins with the acquisition and preprocessing of the Marathi language corpus, MahaCorpus, which comprises an extensive dataset of 24.8 million sentences. To facilitate efficient analysis, we divided this corpus into 20 equal-sized subsets, each containing approximately 1.2 million sentences each. This segmentation approach helped in preserving diversity in text while managing computational complexity.

\subsection{\textbf{TF-IDF (Term Frequency-Inverse Document Frequency)}}

The Term Frequency-Inverse Document Frequency (TF-IDF) approach is a fundamental technique in Natural Language Processing (NLP) used to evaluate the importance of words within a given corpus. TF-IDF quantifies the relevance of a term in a document by considering its frequency in that document and inversely scaling it based on the term's prevalence across the entire corpus. This method enables the identification of key terms that are distinctive to specific documents, contributing significantly to tasks such as text analysis, information retrieval, and document classification.
To implement the TF-IDF approach, we calculated the following metrics for each term in the 20 subsets of the Marathi corpus:
\vspace{8pt}

\begin{figure*}[!ht]
    \begin{minipage}{0.45\textwidth}
    \centering
    \includegraphics[width=\linewidth]{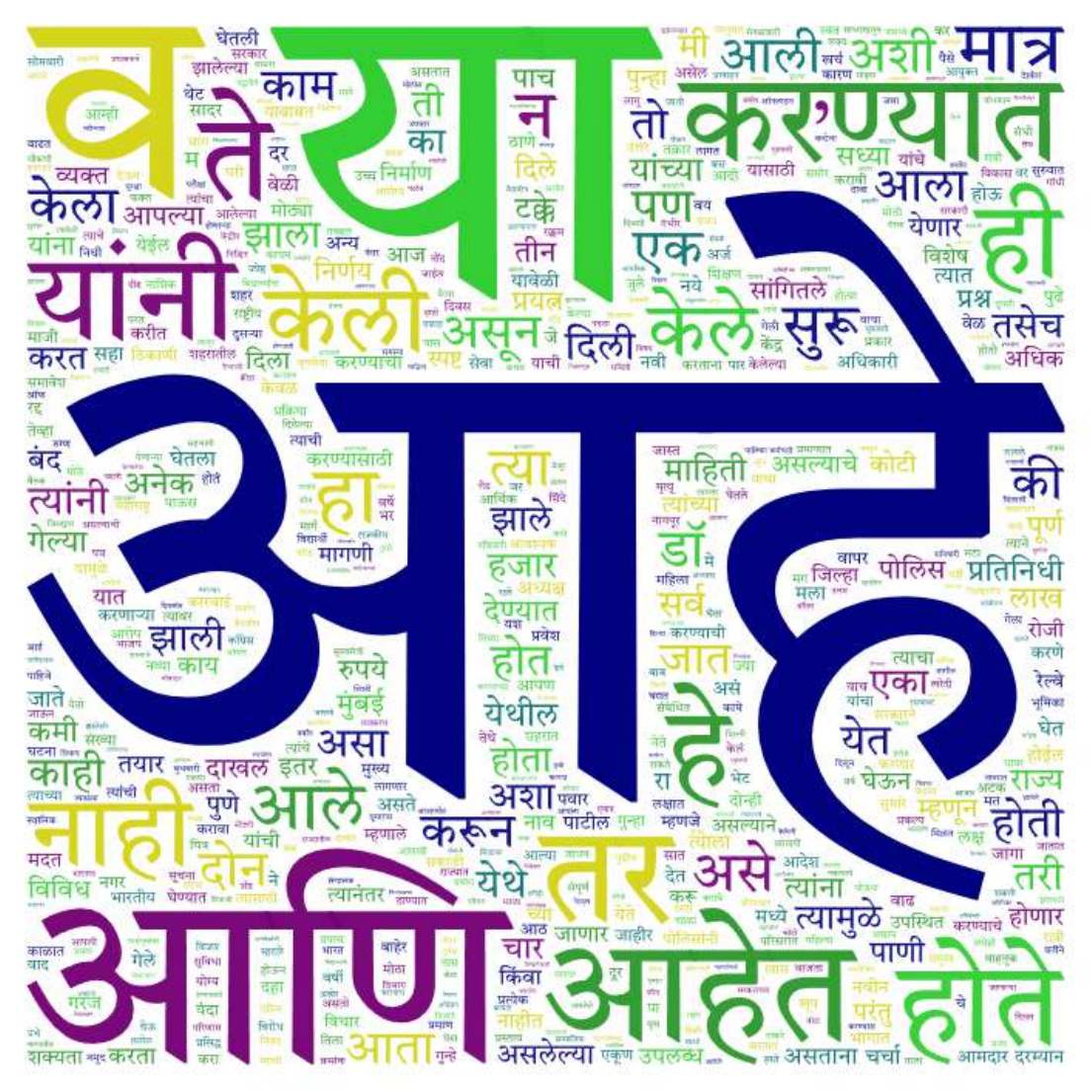}
    \caption{Word Cloud for the top 5k words with lowest TF-IDF scores for all chunks combined}
    \label{Word Cloud}
  \end{minipage}
  \hfill
  \begin{minipage}{0.50\textwidth}
    \centering
    \includegraphics[width=\linewidth]{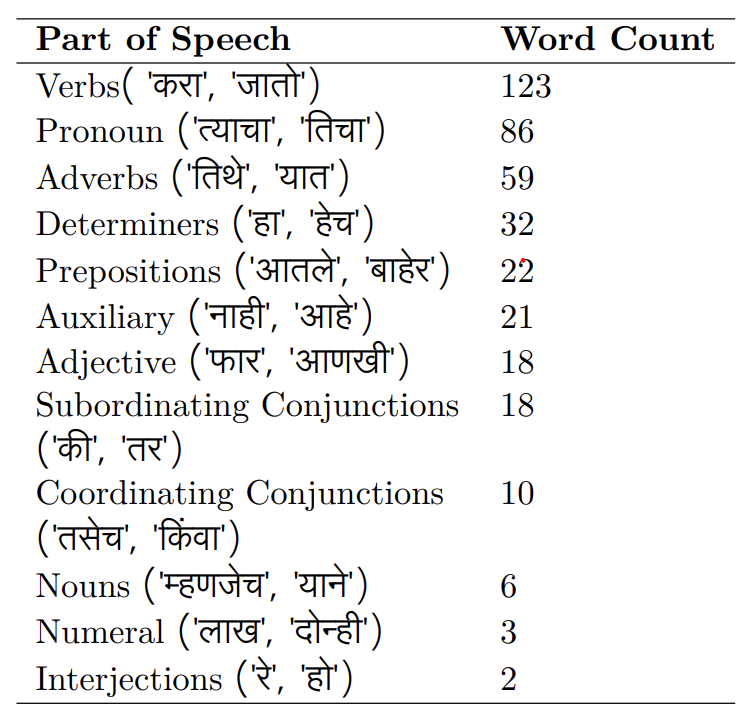}
    \caption{Stopwords Categorized by POS tags}
    \label{citation-guide}
  \end{minipage}%
\end{figure*}

\begin{enumerate}
    \item \textbf{Term Frequency (TF):} \vspace{8pt}
\\  Term Frequency (TF) is the number of times a particular term occurs in a document divided by the total number of terms in that document. It highlights the importance of a term within a specific document.
\vspace{8pt}
    \begin{equation}
    {\textbf{TF}(t, d) = \frac{\small \textbf{Frequency of t in document d}}{\small \textbf{Total terms in document d}}}
    \end{equation}
    \vspace{8pt}
    \item \textbf{Document Frequency (DF):} \vspace{8pt}
\\ Document frequency is calculated by the number of documents containing a particular term divided by the total number of documents in the corpus. It is a measure of how often the appearance of a particular term across a collection of documents.
\vspace{8pt}
    \begin{equation}
    \textbf{DF}(t) = \frac{\small \textbf{No. of documents containing t}}{\small \textbf{Total no. of documents in corpus}}
    \end{equation}
    \vspace{8pt}
    \item \textbf{Inverse Document Frequency (IDF):} \vspace{8pt}
\\ Inverse Document Frequency (IDF) is calculated by taking the logarithm of the reciprocal of the Document Frequency (DF). It demonstrates the rarity or uniqueness of the term in the corpus.
    \begin{equation}
    \large{\textbf{IDF}(t) = \log\left(\frac{1}{\textbf{DF}(t)}\right)}
    \end{equation}
\end{enumerate}

The process of calculating the TF-IDF scores for the words in the corpus is as follows \cite{fayaza2021towards}:
\begin{itemize}
    \item Term Frequency(TF) is calculated for each term in the document. 
    \vspace{4pt}
    \item Each term’s Document Frequency(DF) is calculated.
    \vspace{4pt}
    \item The Inverse Document Frequency(IDF) for each term is calculated.
    \vspace{4pt}
    \item The TF*IDF value is calculated for each term 
    \vspace{4pt}
    \item List TF*IDF values in ascending order.
    \vspace{4pt}
    \item Extract the top 5000 words with the lowest TF-IDF values.
    \vspace{4pt}
    \item Repeat the above operations for all 20 chunks of the corpus. 
    \vspace{4pt}
\end{itemize}
\vspace{8pt}

For each term, TF-IDF is calculated as the product of TF and IDF. This results in a TF-IDF score that reflects the importance of a term within a subset while considering its uniqueness in the broader corpus. Consequently, each subset yields a list of terms ranked by their TF-IDF scores(Fig. 2).
\subsection{\textbf{Common Word Extraction}}
\vspace{8pt}

To compile a consolidated list of stopwords resonating with the core characteristics of the Marathi language, we performed an intersection operation on the 20 candidate stopword lists. By selecting words consistently prevalent among the top 5000 lowest TF-IDF values across all subsets, we aimed to ensure broader linguistic significance. This pragmatic approach minimizes the impact of subset-specific variations, enhancing the overall relevance and applicability of the curated stopword list. The rationale behind this intersection is rooted in the belief that words consistently appearing across diverse subsets are likely to be frequent in the Marathi language, regardless of specific domain or topic considerations. The details of this process are visually represented in the accompanying flowchart (Fig. 1), providing a clear overview of our criteria for identifying common stopwords.
\vspace{8pt}

\subsection{\textbf{Meticulous Human Evaluation}}
\vspace{8pt}

To refine our stopword list further, we conducted a meticulous evaluation involving three native Marathi speakers, all fluent in the language. The evaluation process employed a comprehensive approach where each word among the 2297 candidate stopwords was analyzed for its semantic value within the context of a sentence. Specifically, we aimed to assess the contribution of each stopword to the overall meaning of the sentence by analysing a sentence containing that specific word and assessing the semantics of the sentence on removing the word. The decision-making process involved a voting mechanism, with each reviewer providing their input on whether a word should be classified as a stopword or not. If two or more reviewers agreed that the removal of a stopword significantly altered the meaning of the sentence, indicating its importance, the word was classified as non-trivial and not included in the stopword list. Conversely, if there was disagreement or if the removal had minimal impact on the sentence's meaning, signifying the term's lack of meaningful contribution, the word was classified as a candidate stopword. This democratic approach ensured that the final set of 400 stopwords was not only linguistically relevant but also aligned with the characteristics of Marathi stopwords. The meticulous nature of this evaluation, coupled with the collective expertise of the reviewers, underscores our commitment to producing a curated stopword list that accurately reflects the linguistic nuances of the Marathi language.
\vspace{8pt}

\section{\textbf{\Large Results and Discussion}}

\vspace{8pt}
Our investigation delved into the intricate dynamics of stopword removal on both long-document classification and sentiment analysis tasks in Marathi language processing. To curate our stopwords list, we initiated the process by selecting the 5,000 least frequent words from each of the 20 segments in our corpus. From this initial pool, a refined list of 2297 common terms emerged, which underwent meticulous manual evaluation to produce a final list of 400 stopwords. Fig. 3 illustrates the categorization of these stopwords into their respective parts of speech, adding an extra layer of linguistic insight to our curated resource.

\begin{center}
    \begin{table}[!ht]
    \centering
    \renewcommand{\arraystretch}{1.75}
    \begin{tabular}{|c|c|}
    \hline
    \textbf{Model} & \textbf{Accuracy} \\
    \hline
    IndicBERT (Stopwords) & 94.24\% \\
    \hline
    MahaBERT (No Stopwords) & 94.02\% \\
    \hline
    MahaBERT (Stopwords) & 94.06\% \\
    \hline
    IndicBERT (No Stopwords) & 93.77\% \\
    \hline
    \addlinespace
    \end{tabular}
    \vspace{2pt}
    \caption{Results of Stopword Removal on Pre-Trained Transformer Models}
    \label{tab:stopwords}
    \end{table}
\end{center}
For testing, we utilized the MahaNews LDC dataset 
by L3Cube \cite{joshi2022l3cube_mahanlp}, a comprehensive long-document classification dataset comprising news articles and headlines. This dataset, which consists of 22,017 training samples, 2,761 testing samples, and 2,750 validation samples, represents a real-world, domain-specific context for our experiments. The results are shown in Table \ref{tab:stopwords}.
\vspace{8pt}

Our initial experiments on long-document classification using pre-trained transformer models, IndicBERT and MahaBERT, revealed impressive accuracies of 94.24\% and 94.02\%, respectively. This underscores the inherent effectiveness of these models on our chosen task. Intriguingly, the removal of stopwords resulted in only a marginal accuracy drop, with MahaBERT maintaining a robust 94.06\% accuracy, and IndicBERT displaying 93.77\%. These findings align with expectations for BERT-based models, pre-trained on full texts \cite{miyajiwala2022sensitivity}, emphasizing the resilience of these models to stopword variations.
\vspace{8pt}

The minor reduction in model accuracy (Table \ref{tab:stopwords}), is consistent with the broader literature on stopwords, and their impact on NLP tasks and it
underscores the significance of crafting language-specific stopword lists while effectively addressing the unique linguistic characteristics of Marathi. Our approach effectively addresses the unique linguistic characteristics of Marathi, offering practical solutions for diverse text processing needs and maintaining the model's effectiveness in various applications.
\vspace{8pt}

\begin{table}[!ht]
    \centering
    \renewcommand{\arraystretch}{1.75}
    \begin{tabular}{|c|c|}
        \hline
        \textbf{Model} & \textbf{Accuracy} \\
        \hline
        MahaBERT (Stopwords present) & 83.11\% \\
        \hline
        MahaBERT (Stopwords removed) & 83.68\% \\
        \hline
        \addlinespace
    \end{tabular}
    \vspace{2pt}
    \caption{Results of Stopword Removal on MahaBERT for Sentiment Analysis}
    \label{tab:sentiment}
\end{table}

The insights gained from long-document classification were further extended to sentiment analysis, a task critical for understanding user opinions and emotions in Marathi text. Leveraging the MahaBERT model on the MahaSENT dataset, specifically curated for Marathi sentiment analysis, we observed a noteworthy improvement in accuracy. The initial accuracy of MahaBERT without stopword removal stood at 83.11\%, and post-stopword removal, the accuracy surged to 83.68\%. This improvement highlights the impact of stopwords on different NLP applications and underscores the practical benefits of our carefully curated stopwords list.The results of which are shown in Table. \ref{tab:sentiment}.
\vspace{8pt}

It is worth noting that our work represents a pioneering effort in the curation of stopwords for the Marathi language, with a comprehensive and methodical approach leading to a list of 400 stopwords. This list, categorized by parts of speech, serves as a foundational resource for future Marathi NLP research, providing not only efficiency gains but also a nuanced understanding of linguistic characteristics. No prior work has addressed Marathi stopwords with such depth and scope, marking our study as a significant contribution to the growing field of Marathi language processing.
\vspace{8pt}

\section{\textbf{\Large Conclusion}}

In conclusion, our research makes a substantial contribution to the field of Marathi natural language processing by introducing a meticulously curated list of 400 stopwords tailored specifically to the nuances of the Marathi language. The extensive nature of this list, derived through a rigorous process of selecting the 5,000 least frequent words from each of the 20 segments in our corpus and subsequent manual curation, distinguishes our work as one of the first comprehensive endeavours in the curation of Marathi stopwords. The careful consideration of linguistic nuances, as evident in categorising stopwords into their respective parts of speech, provides a resource that significantly enhances text processing efficiency for various applications and research domains.
\vspace{8pt}

Our study, encompassing long-document classification and sentiment analysis tasks, yielded insightful results. In long-document classification, where we employed pre-trained transformer models IndicBERT and MahaBERT, the observed minor reduction in accuracy post-stopword removal aligns with the broader literature on stopwords, particularly for BERT-based models pre-trained on full texts \cite{miyajiwala2022sensitivity}. This work serves as a foundational exploration into the impact of stopwords on NLP tasks in the Marathi language.
\vspace{8pt}

Furthermore, our foray into sentiment analysis using the MahaBERT model showcased a notable improvement in accuracy from 83.11\% to 83.68\% after the removal of stopwords. This dual-task analysis underscores the nuanced and varied impact of stopwords across different NLP applications.
\vspace{8pt}

Our ongoing efforts will be dedicated to further refining this curated resource, with a keen focus on expanding its applications across various Marathi language processing tasks. This research lays the groundwork for the development of language-specific stopword lists, offering practical solutions for diverse text processing needs and reinforcing the importance of linguistic considerations in NLP research and applications.

\section{\textbf{\Large Acknowledgments}}
\vspace{8pt}

This work was done under the L3Cube Labs, Pune mentorship program. We would like to express our gratitude towards our mentors at L3Cube for their continuous support and encouragement. This work is a part of the L3Cube-MahaNLP project \cite{joshi2022l3cube_mahanlp}.
\vspace{8pt}

\bibliography{main}
\bibliographystyle{plain}
\end{document}